\documentclass[sigconf,nonacm,]{acmart}

\usepackage{amsmath}
\usepackage{algorithmic}
\usepackage{textcomp}
\usepackage[table]{xcolor}
\usepackage{balance}
\usepackage[utf8]{inputenc}
\usepackage[T1]{fontenc}
\usepackage[caption=false,font=footnotesize]{subfig}
\usepackage{booktabs}
\usepackage{nicefrac}
\usepackage{microtype}
\usepackage{placeins}
\usepackage{tikz}
\usepackage{multirow}
\usepackage{array}
\usepackage{hyperref}
\usepackage{graphicx}
\usepackage{arydshln}

\usetikzlibrary{arrows.meta, positioning, fit}

\def\BibTeX{{\rm B\kern-.05em{\sc i\kern-.025em b}\kern-.08em
T\kern-.1667em\lower.7ex\hbox{E}\kern-.125emX}}

\setcopyright{none}
\renewcommand\footnotetextcopyrightpermission[1]{}

\begin{document}

\title{Can an Actor-Critic Optimization Framework Improve Analog Design?}

\author{%
Sounak Dutta\textsuperscript{*},
Fin Amin\textsuperscript{*},
Sushil Panda,
Jonathan Rabe,
Yuejiang Wen,
and Paul D. Franzon
}

\affiliation{%
  \institution{North Carolina State University}
  \department{Department of Electrical and Computer Engineering}
  \city{Raleigh}
  \state{North Carolina}
  \country{USA}
}

\email{{sdutta6, samin2, spanda4, jcrabe, wyuejia, paulf}@ncsu.edu}

\renewcommand{\shortauthors}{Dutta, Amin, et al.}

\begin{abstract}

Analog design often slows down because even small changes to device sizes or biases require expensive simulation cycles, and high-quality solutions typically occupy only a narrow part of a very large search space. While existing optimizers reduce some of this burden, they largely operate without the kind of judgment designers use when deciding where to search next. This paper presents an actor-critic optimization framework (ACOF) for analog sizing that brings that form of guidance into the loop. Rather than treating optimization as a purely black-box search problem, ACOF separates the roles of proposal and evaluation: an actor suggests promising regions of the design space, while a critic reviews those choices, enforces design legality, and redirects the search when progress is hampered. Across our test circuits, ACOF improves the top-10 figure of merit by an average of 38.9\% over the strongest competing baseline and reduces regret by an average of 24.7\%, with peak gains of 70.5\% in FoM and 42.2\% lower regret on individual circuits. By combining iterative reasoning with simulation-driven search, the framework offers a more transparent path toward automated analog sizing across challenging design spaces.

\end{abstract}


\maketitle

\begingroup
\renewcommand{\thefootnote}{*}
\footnotetext{Equal contribution \\ An earlier version of this manuscript was accepted for presentation at the non-archival NSF Workshop on Agents for Chip Design Automation (AI4EDA '26). }
\endgroup

\pagestyle{plain}
\thispagestyle{plain}

\section{Introduction and Motivation} \label{sec:intro}


The demands on modern analog design continue to increase with new technology nodes, tighter specifications, and faster development cycles. Each node changes transistor behavior, invalidates earlier biasing strategies, and requires renewed exploration of the design space. From an optimization standpoint, this is no longer merely schematic tuning, but the search for a feasible region where gain, bandwidth, phase margin, and power can be satisfied together. Because performance varies nonlinearly with sizing, designers need SPICE simulations, and more knobs mean more costly trials. Even after the topology is fixed, analog sizing is rarely straightforward. It typically requires repeated evaluation and careful adjustment to satisfy several coupled specifications while managing trade-offs among performance, power, and area. In practice, analog design is still a slow and labor-intensive process, requiring extensive simulation and considerable designer expertise \cite{barros2010analog, ahmadzadeh2024using, ahmadzadeh2025anacraft}.

\begin{figure}[ht] 
\centering 
 \includegraphics[width=0.83\linewidth, trim=20mm 30mm 20mm 0mm, clip]{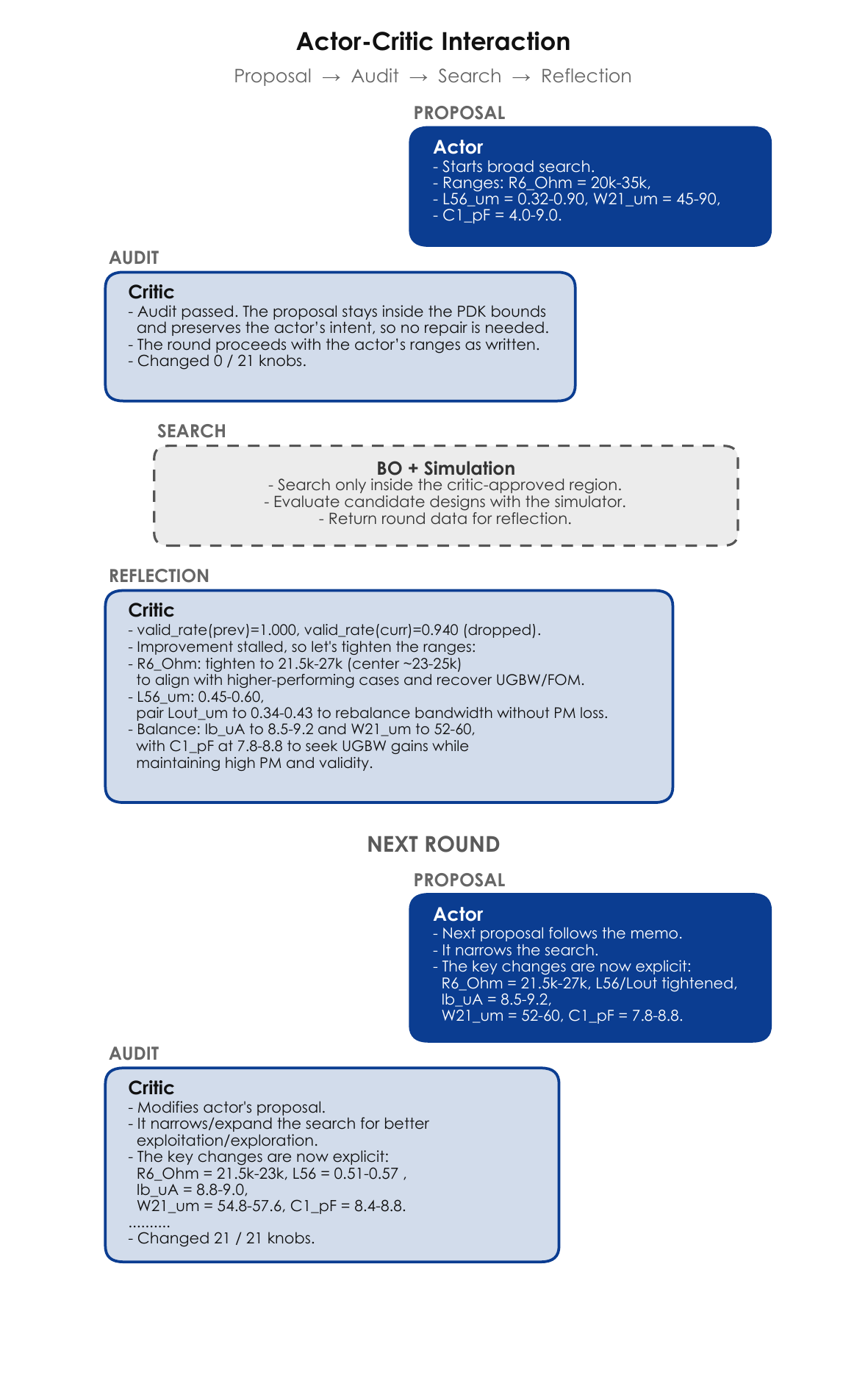}  
%
\caption{Example of interactions between our actor and critic during analog sizing optimization. Our framework defines a closed-loop optimization process that alternates between proposal, audit, search, and reflection.} \label{fig:convo} 
\end{figure}


\begin{figure*}[ht]
    \centering 
    \includegraphics[width=0.7\linewidth, trim=2mm 20mm 2mm 37mm, clip]{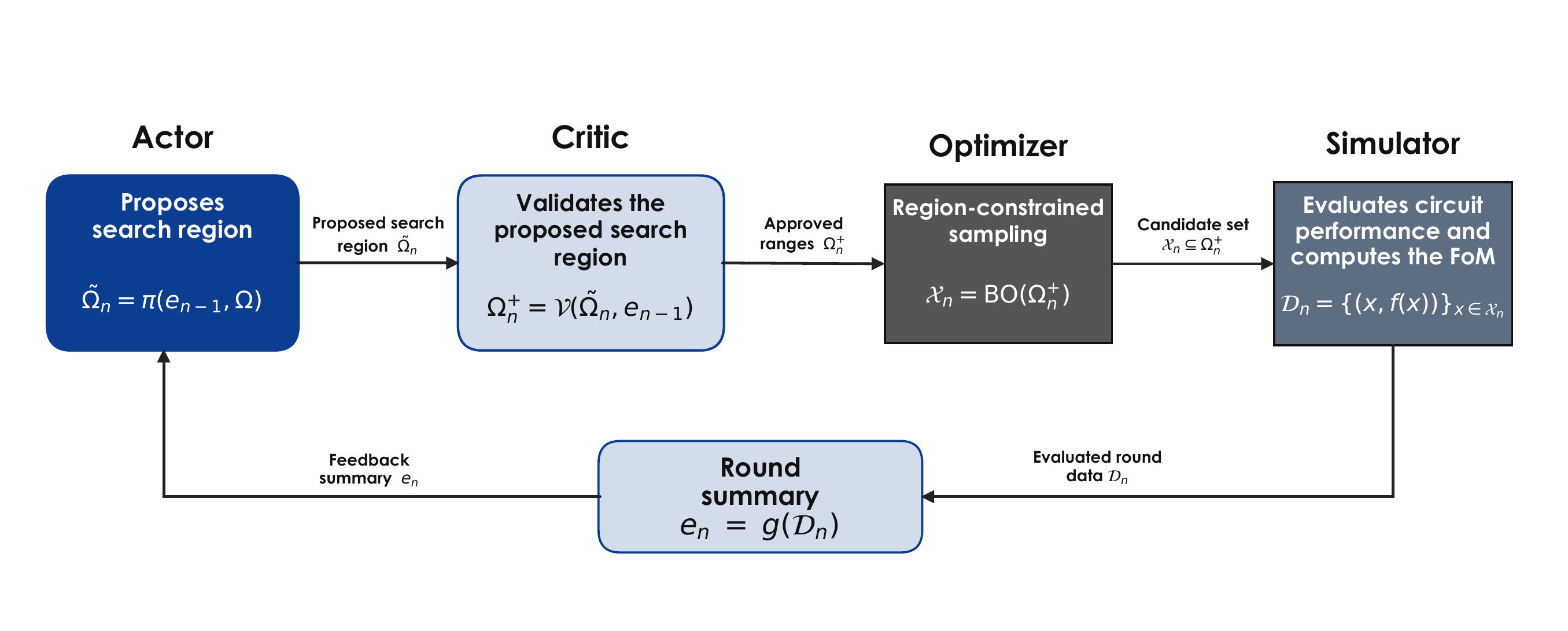}
    \caption{An overview of our actor–critic optimization framework (ACOF). At each time step, $n$, the actor proposes a candidate search region $\tilde{\Omega}_n$, the critic adjusts it to a PDK-legal region $\Omega_n^{+}$, BO selects candidates $x \in \Omega_n^{+}$ for simulation, and simulator outcomes are summarized into $e_n$ for the next round. ACOF is agnostic to the specific choice of BO.}
    \label{fig:Acof_overview}
\end{figure*}

For that reason, analog sizing has long been viewed as a constrained optimization problem. From that perspective, the central challenge is not simply to evaluate better candidate points, but to direct the search toward regions of the design space that are more likely to produce feasible and high-performing solutions under competing design constraints \cite{lyu2017efficient, touloupas2021local, ahmadzadeh2024using, ahmadzadeh2025anacraft}. Human designers address this problem differently. They do not treat the search space as uniformly promising; they use judgment to focus attention, interpret trade-offs, and revise direction when progress stalls.

This paper is motivated by the idea that such guidance should become part of the optimization loop itself. Recent LLM-based studies in electronic design suggest that language models can contribute to design automation as reasoning agents rather than as mere text generators \cite{liu2025layoutcopilot, chen2024llm, liu2024ampagent, lai2025analogcoder, yin2024ado, lorakd}. Building on that perspective, \textbf{our contribution} is to investigate casting analog sizing as an actor–critic process in which one agent proposes promising regions of the design space, another evaluates whether the search direction should be preserved or revised, and a conventional optimizer explores within the resulting region. In this way, automated analog sizing can become not only more sample-efficient, but also more interpretable and more aligned with the way experienced designers reason about difficult trade-offs.

\section{Prior Art} \label{sec:prior_art}
\subsection{BO for Analog Circuit Sizing}

Bayesian optimization (BO) has advanced analog circuit sizing in several important ways \cite{wen2022high}. WEIBO \cite{lyu2017efficient} formulates sizing as constrained BO with Gaussian-process (GP) surrogates and weighted expected improvement, and extends the approach to multi-objective design. Local BO \cite{touloupas2021local} scales this idea to larger spaces through trust-region-based local models and batch candidate evaluation. tSS-BO \cite{gu2024tss} targets the high-dimensional setting by restricting search to an effective truncated subspace while retaining local GP guidance and Bayesian selection. Despite these advances, BO methods for analog sizing still rely on surrogates trained only on evaluations from the target circuit, so each new circuit or technology node largely requires restarting the search from scratch.

\subsection{Language Driven Optimization}
Recent LLM-based work in analog design has begun to open several distinct directions rather than one unified path. LayoutCopilot \cite{liu2025layoutcopilot} reimagines analog layout interaction by turning natural-language design requests into executable layout actions through a multi-agent framework, while LLANA \cite{chen2024llm} brings language models into the BO loop to generate design-dependent layout constraints, particularly net-weighting parameters for layout synthesis. AmpAgent \cite{liu2024ampagent} approaches amplifier design as a coordinated reasoning process across literature understanding, formula derivation, and device sizing. In contrast, AnalogCoder \cite{lai2025analogcoder} treats analog design as a code-generation problem, using a training-free workflow to produce and iteratively correct Python-based circuit implementations. ADO-LLM \cite{yin2024ado} more directly couples an LLM with Bayesian optimization for circuit sizing, using the model to propose promising design candidates alongside a GP-based search process. LEDRO \cite{kochar2025ledro}, in turn, shifts the emphasis from asking the LLM for a single sizing answer to asking it to progressively narrow the search region so that optimization can proceed within a more meaningful part of the design space. More recent frameworks extend this picture in complementary ways: EEsizer \cite{liu2025eesizer} formulates transistor sizing as a closed-loop LLM-guided process built around simulation, analysis, and iterative parameter updates, while LLM-USO \cite{somayaji2025llm} focuses on structured knowledge reuse across related circuits.

\subsection{Takeaways from Prior Art}

These prior works point to the same recurring pressures: GP-based BO becomes harder to scale as the sample set and parameter dimension grow. Constrained search remains difficult, and exploration becomes increasingly fragile in high-dimensional spaces. Furthermore, these methods mainly act as black boxes \cite{touloupas2021local, chen2024llm, gu2024tss}. Recent LLM-based methods have begun to address these limitations in different ways: some strengthen the reasoning and decomposition side of analog design \cite{liu2024ampagent}, some couple LLMs with BO to improve candidate generation within the optimization loop \cite{yin2024ado} and some demonstrate the broader value of self-reflection \cite{shinn2023reflexion} for iterative improvement in optimization workflows \cite{kochar2025ledro}. But in these systems, guidance is still driven mainly by simulator feedback or by the model’s own reflective process, leaving a natural opening for an independent critic that can examine a proposal from outside the actor’s reasoning stream. That distinction matters because prompt-heavy LLM systems already report performance dilution under long, multitask context \cite{liu2025layoutcopilot}. In relation to this, long-context retrieval studies show that current LLMs can struggle even on simple fact-retrieval tasks as context grows \cite{nelson2024needle}. Our method is designed around exactly this gap. Considering these takeaways, we introduce ACOF in the following section.
\begin{table*}[ht]
\caption{Results across the four test circuits. Each circuit forms one multi-row block at left. Values are mean with standard errors in subscripts. Bold indicates the best baseline method within each circuit and metric. Qwen was used as the LLM.}
\label{tab:qwen_super_table}
\centering
\setlength{\tabcolsep}{3pt}
\renewcommand{\arraystretch}{1.25}
\resizebox{\textwidth}{!}{%
\begin{tabular}{c l c c c c >{\columncolor{gray!25}}c c >{\columncolor{gray!25}}c c >{\columncolor{gray!25}}c}
\toprule
\multicolumn{2}{c}{} & \multicolumn{5}{c}{Top-10 Design Quality} & \multicolumn{2}{c}{Reliability (\%)} & \multicolumn{2}{c}{Exploration/Exploitation} \\
\cmidrule(lr){3-7}\cmidrule(lr){8-9}\cmidrule(lr){10-11}
  & Method & Gain (dB) $\uparrow$ & UGBW (MHz) $\uparrow$ & PM (deg) $\uparrow$ & Power (mW) $\downarrow$ & FoM $\uparrow$ & Sim. Valid $\uparrow$ & Phys. Feasible $\uparrow$ & Regions $\uparrow$ & Regret $\downarrow$ \\
\midrule
\multirow{5}{*}{\rotatebox[origin=c]{90}{\shortstack[c]{130nm\\12 Params}}} & Target Spec. & 85.00 & 900.00 & 100.00 & 0.554 & 0.00 & - & - & - & - \\
\cdashline{2-11}[0.4pt/1.2pt]
 & ACOF & \textbf{83.65$_{0.64}$} & \textbf{785.30$_{39.71}$} & 101.35$_{1.13}$ & 0.658$_{0.045}$ & \textbf{-0.19$_{0.01}$} & \textbf{97.0$_{0.1}$} & \textbf{86.2$_{1.4}$} & 2.0$_{0.0}$ & \textbf{0.25$_{0.01}$} \\
 & LEDRO & 83.15$_{0.55}$ & 717.63$_{45.51}$ & 111.89$_{0.74}$ & 0.657$_{0.029}$ & -0.24$_{0.02}$ & 95.4$_{0.9}$ & 78.9$_{1.8}$ & 2.7$_{0.3}$ & 0.30$_{0.01}$ \\
 & Pure-BO & 76.24$_{1.45}$ & 673.92$_{22.65}$ & \textbf{128.23$_{3.27}$} & \textbf{0.587$_{0.014}$} & -0.39$_{0.02}$ & 81.0$_{0.3}$ & 26.9$_{0.1}$ & \textbf{3.0$_{0.8}$} & 0.37$_{0.02}$ \\
 & Human Expert & 72.77 & 851.48 & 173.59 & 0.821 & -0.45 & - & - & - & - \\
\midrule
\multirow{5}{*}{\rotatebox[origin=c]{90}{\shortstack[c]{180nm\\12 Params}}} & Target Spec. & 95.00 & 500.00 & 100.00 & 0.194 & 0.00 & - & - & - & - \\
\cdashline{2-11}[0.4pt/1.2pt]
 & ACOF & \textbf{90.99$_{0.55}$} & 195.59$_{10.25}$ & 119.77$_{2.23}$ & \textbf{0.200$_{0.001}$} & \textbf{-0.52$_{0.02}$} & \textbf{100.0$_{0.0}$} & \textbf{90.2$_{1.9}$} & \textbf{5.0$_{1.2}$} & \textbf{0.59$_{0.02}$} \\
 & LEDRO & 90.88$_{0.20}$ & 240.69$_{78.42}$ & 117.72$_{1.86}$ & 0.324$_{0.104}$ & -0.64$_{0.01}$ & 100.0$_{0.0}$ & 84.9$_{1.0}$ & 2.0$_{0.0}$ & 0.64$_{0.01}$ \\
 & Pure-BO & 86.79$_{0.29}$ & \textbf{253.71$_{7.97}$} & \textbf{127.97$_{1.23}$} & 0.359$_{0.024}$ & -0.74$_{0.02}$ & 96.1$_{0.5}$ & 36.1$_{2.6}$ & 2.0$_{0.0}$ & 0.69$_{0.04}$ \\
 & Human Expert & 81.62 & 365.69 & 127.42 & 0.361 & -0.68 & - & - & - & - \\
\midrule
\multirow{5}{*}{\rotatebox[origin=c]{90}{\shortstack[c]{130nm\\17 Params}}} & Target Spec. & 85.00 & 20.00 & 100.00 & 0.277 & 0.00 & - & - & - & - \\
\cdashline{2-11}[0.4pt/1.2pt]
 & ACOF & \textbf{88.56$_{2.77}$} & 29.45$_{7.68}$ & \textbf{104.59$_{22.57}$} & \textbf{0.290$_{0.008}$} & \textbf{-0.13$_{0.05}$} & \textbf{97.7$_{0.8}$} & 81.7$_{10.3}$ & \textbf{3.7$_{0.7}$} & \textbf{0.26$_{0.02}$} \\
 & LEDRO & 87.06$_{1.53}$ & 20.62$_{1.04}$ & 71.73$_{2.23}$ & 0.475$_{0.141}$ & -0.44$_{0.17}$ & 93.9$_{1.6}$ & \textbf{82.3$_{6.5}$} & 2.7$_{0.5}$ & 0.45$_{0.12}$ \\
 & Pure-BO & 85.69$_{0.40}$ & \textbf{50.49$_{15.61}$} & 74.93$_{5.47}$ & 0.656$_{0.100}$ & -0.73$_{0.01}$ & 84.6$_{0.6}$ & 68.4$_{0.6}$ & 1.3$_{0.5}$ & 0.65$_{0.03}$ \\
 & Human Expert & 78.33 & 9.93 & 80.06 & 0.560 & -0.91 & - & - & - & - \\
\midrule
\multirow{5}{*}{\rotatebox[origin=c]{90}{\shortstack[c]{180nm\\21 Params}}} & Target Spec. & 85.00 & 1200.00 & 120.00 & 1.320 & 0.00 & - & - & - & - \\
\cdashline{2-11}[0.4pt/1.2pt]
 & ACOF & 94.57$_{2.38}$ & \textbf{1095.36$_{18.60}$} & 84.20$_{3.69}$ & 1.329$_{0.014}$ & \textbf{-0.24$_{0.02}$} & 99.3$_{0.3}$ & \textbf{85.6$_{2.4}$} & \textbf{2.7$_{0.3}$} & \textbf{0.32$_{0.01}$} \\
 & LEDRO & \textbf{102.43$_{4.70}$} & 759.88$_{178.60}$ & 90.91$_{10.95}$ & \textbf{1.232$_{0.090}$} & -0.44$_{0.07}$ & \textbf{99.7$_{0.2}$} & 74.6$_{3.3}$ & 1.0$_{0.8}$ & 0.47$_{0.05}$ \\
 & Pure-BO & 98.06$_{2.13}$ & 469.33$_{59.13}$ & \textbf{123.24$_{7.15}$} & 1.243$_{0.068}$ & -0.61$_{0.01}$ & 91.4$_{0.1}$ & 19.3$_{1.0}$ & 1.7$_{0.7}$ & 0.58$_{0.01}$ \\
 & Human Expert & 72.69 & 1138.87 & 124.78 & 2.210 & -0.51 & - & - & - & - \\
\bottomrule
\end{tabular}%
}
\end{table*}

\section{Our Approach} \label{sec:approach}
\subsection{Why Use an Actor-Critic Structure?}

The distinction between policy-based and value-based methods provides a useful lens for understanding why an actor--critic structure is a natural design choice in optimization. In reinforcement learning (RL), policy methods learn a direct mapping from states to actions, whereas value-based methods learn an estimate of expected return to guide action selection, as in REINFORCE/TRPO versus Q-learning/DQN \cite{sutton2018reinforcement,sutton2000policy,mnih2015human,schulman2015trpo}. The historical trend was not a simple shift from policy learning to value learning, since value-based deep RL was already central in the mid-2010s through DQN. A more meaningful shift was toward methods in which action proposals are shaped by an explicit evaluator. Actor-critic methods made this especially clear by pairing a policy with a learned critic that scores actions or states and stabilizes learning through lower-variance updates \cite{sutton2000policy,haarnoja2018sac}. More broadly, this points to a lesson larger than any one RL algorithm: many successful learning systems benefit when generation and judgment need not share the same mechanism. Methods such as Soft Actor-Critic, TD3, and MuZero push this logic further by placing learned value estimates, critics, and planning modules at the center of decision making \cite{haarnoja2018sac, TD3, schrittwieser2020muzero}. In parallel, reward-learning and preference-based frameworks, including RLHF, further reinforce the idea that learning what should evaluate behavior can be as important as learning the behavior itself \cite{christiano2017preferences,ouyang2022instructgpt}.

\begin{figure*}[!ht]
    \centering 
    \includegraphics[width=0.82\linewidth, trim=2mm 2mm 2mm 2mm, clip]{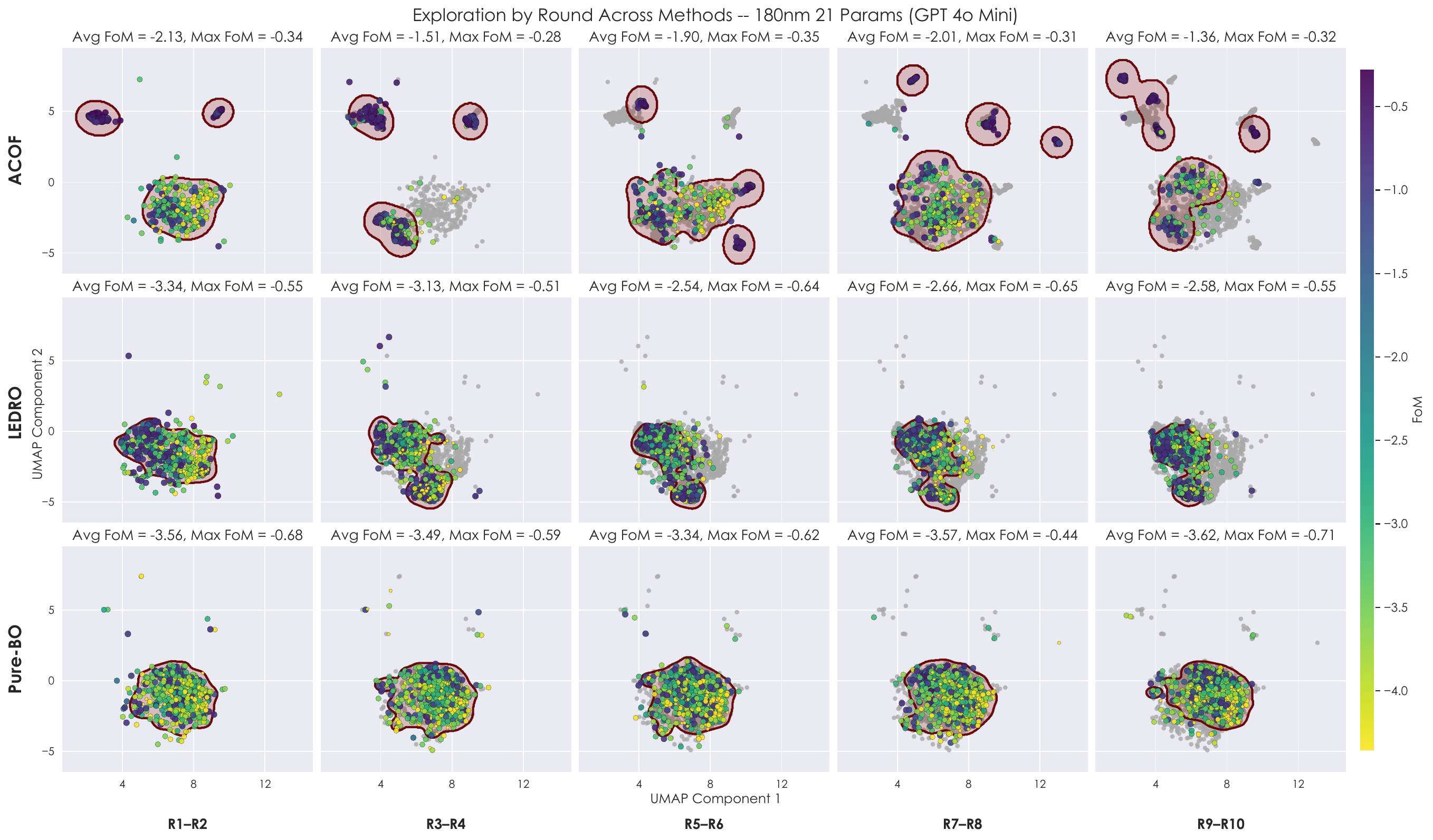}
    \caption{UMAP projections of the sizing parameters illustrate how the design space was explored over optimization rounds. To aid visualization, we outlined regions being explored in the current rounds in red and rendered the cumulative explored designs from past rounds in gray. ACOF judiciously explores the design space by targeting regions corresponding to high performance--as indicated by the higher round-wise Avg and Max FoM. Our technique successfully explored the regions above $\texttt{UMAP Component 2} > 1$. On the other hand, the baselines were apprehensive of exploration and achieved worse FoMs.}
    \label{fig:umap_gpt4mini}
\end{figure*}

Our approach is motivated by the same principle, but applied to analog sizing rather than sequential control. Our framework is not an RL algorithm in the standard sense; instead, it adopts the structural lesson behind actor-critic methods by separating the proposal of a search region from its evaluation. In analog optimization, this separation is useful because a proposed region may be promising in spirit yet still be ill-posed, physically implausible, or wasteful under a fixed simulation budget. An independent critic therefore serves as an explicit evaluative mechanism that helps stabilize and redirect the search before expensive simulation is spent. 

\subsection{The Actor-Critic Optimization Framework}\label{sec:acof_method}

We formulate analog circuit sizing as $x^\star = \arg\max_{x \in \Omega} f(x)$, where \(x\) denotes the vector of tunable circuit parameters, \(\Omega\) denotes the global feasible design domain induced by technology bounds and hard circuit constraints, which is primarily defined by the PDK and \(f(x)\) denotes the scalar figure of merit (FoM) obtained from a costly circuit simulation.

Our approach, the actor-critic optimization framework (ACOF), is organized as a role-separated reasoning loop with two logical agents: an \emph{actor}, which proposes where the search should move next, and a \emph{critic}, which reviews that proposal before optimization proceeds. This structure turns the familiar design-review pattern of analog practice into an iterative computational procedure.

In our pipeline the actor does not output a single design point; instead, it proposes a search region within which the downstream optimizer should explore. At initialization, the actor is seeded by a small set of valid calibration examples, denoted by \(\mathcal{C}_0\) \cite{kochar2025ledro}, and produces the first candidate region \(\tilde{\Omega}_1 = \pi(\mathcal{C}_0,\Omega)\). After the first round, the actor is updated by a round summary \(e_{n-1}\) that condenses what the previous round revealed, including the best design found so far, representative high-quality samples, and the critic's reflection on that round, so that for later rounds it proposes \(\tilde{\Omega}_n = \pi(e_{n-1},\Omega)\). In the implemented loop, the actor is therefore driven not by the previous corrected region itself, but by the evidence generated from searching within it. Instead of allowing optimization to unfold as a blind sequence of samples, ACOF inserts an explicit stage of judgment between proposal and evaluation. As the search progresses, the critic also produces short reflection memos \cite{shinn2023reflexion} that capture recurring patterns, useful corrections, and promising directions, thereby giving the loop continuity across rounds. An overview is given in Fig.~\ref{fig:Acof_overview}.


Before any BO step is performed, the critic audits the actor's proposal and converts it into the region actually used for search, written as \(\Omega_n^{+} = \mathcal{V}(\tilde{\Omega}_n,e_{n-1})\), where \(\mathcal{V}(\cdot)\) denotes the critic's validation and correction step. Here \(\Omega_n^{+}\) is the post-audit region for round \(n\): it is the actor's proposal after being checked, corrected when necessary, and accepted for optimization. Conceptually, this stage both preserves the actor's intent when that intent is coherent and stabilizes the search when the proposal is malformed, overly narrow, overly loose, or otherwise inconsistent with the feasible domain. In this sense, the critic functions less as a second optimizer than as a gatekeeper that keeps exploration disciplined. 

Once \(\Omega_n^{+}\) has been established, BO and simulation are carried out within that region under a fixed evaluation budget. \(\mathcal{X}_n \sim \mathrm{BO}(\Omega_n^{+})\) denotes the candidate designs selected for evaluation in round \(n\), then simulation produces the round data set \(\mathcal{D}_n = \{(x, f(x))\}_{x \in \mathcal{X}_n}\), where \(f(x)\) is the corresponding figure of merit. BO is therefore not used to search the full design space at every step, but to search efficiently within the region approved for that round.

After simulation, the framework compresses the round's outcomes into a summary \(e_n = g(\mathcal{D}_n)\), where \(g(\cdot)\) denotes the round-level reflection process. This summary captures the practical meaning of the round: which parts of the region were productive, whether progress was made, and what kind of adjustment should guide the next proposal. The critic's memo is part of this summary, so the feedback is not purely numerical but also interpretive. The next round then follows the same pattern, with \(\tilde{\Omega}_{n+1} = \pi(e_n,\Omega)\) and \(\Omega_{n+1}^{+} = \mathcal{V}(\tilde{\Omega}_{n+1},e_n)\).

Taken together, these steps define a closed-loop optimization process that alternates between proposal, audit, search, and reflection (see Fig.~\ref{fig:convo}). The actor proposes where to search, the critic determines whether that proposal is suitable for actual optimization, BO explores within the approved region, and the resulting evidence is distilled into guidance for the next round. This separation between proposing a region and approving a region is the core operational idea of ACOF: it allows the search space to be refined progressively while remaining anchored to the feasible structure of the circuit.

\section{Experiments} \label{sec:experiments}
We evaluate three optimization frameworks for their effectiveness in identifying high-quality circuit designs. In all cases, the objective is to maximize a scalar FoM that captures the target performance requirements. For each circuit, run, and method, we form a run-level summary, and then we report the average and standard error of the run-level summaries across the runs. Our analysis considers three groups of metrics: design quality, reliability, and exploration/exploitation.

\begin{figure}[!h]
\centering 

\subfloat[Two-stage amplifier\label{fig:schematics_a}]{
    \includegraphics[width=0.6\linewidth, trim=15mm 5mm 10mm 10mm, clip]{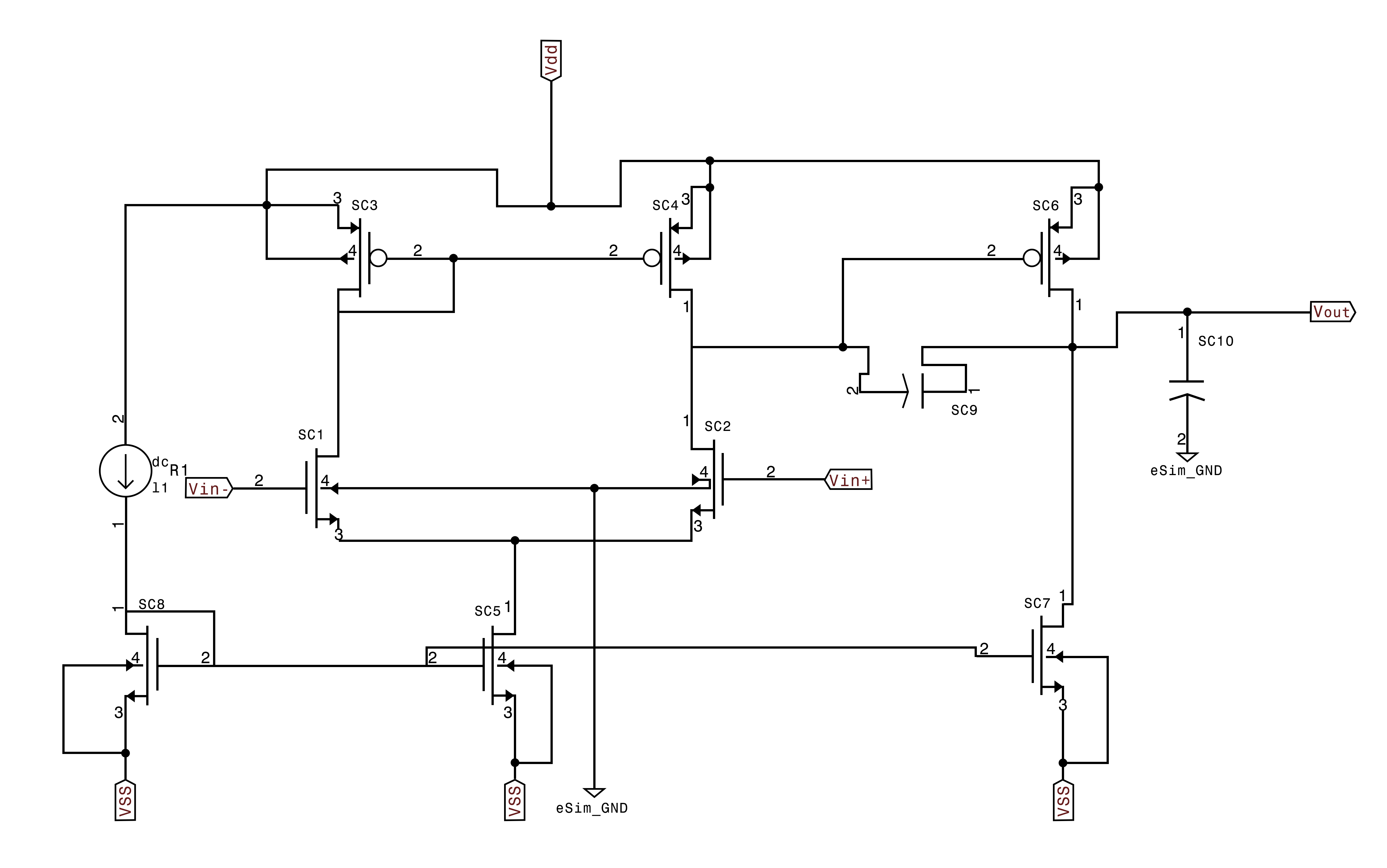}
}

\vspace{2mm}

\subfloat[Folded cascode single-ended amplifier\label{fig:schematics_b}]{
    \hspace*{-5mm}\includegraphics[width=0.825\linewidth, trim=40mm 70mm 40mm 50mm, clip]{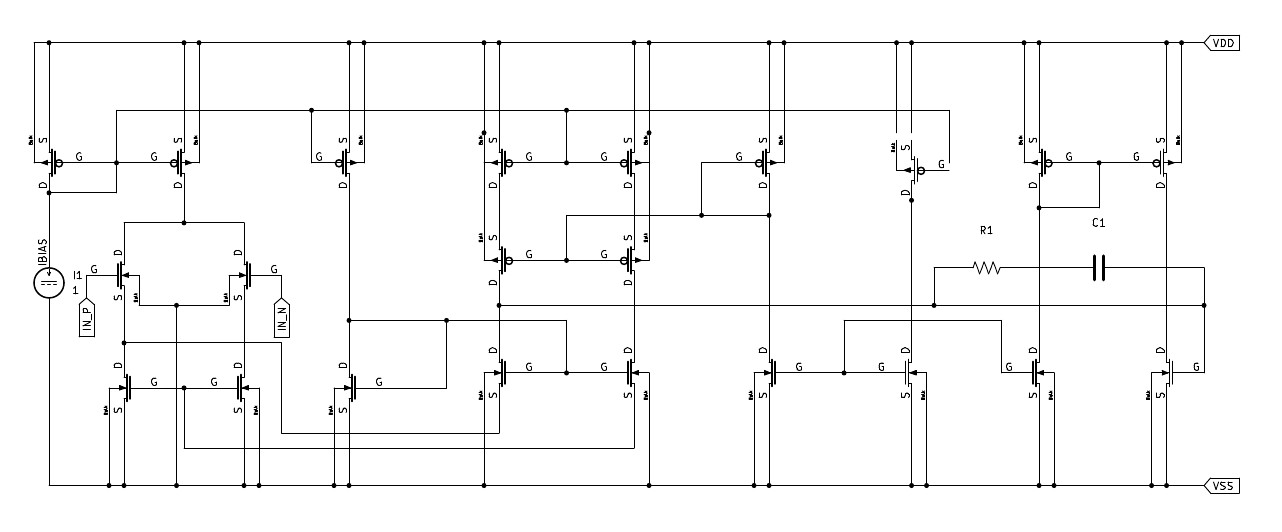}
}

\caption{The circuit schematics of our benchmarks. For an OpAmp, gain, bandwidth, phase margin, and power are coupled design specifications. Higher DC gain can introduce additional poles; this affects phase margin. Also, improving UGBW usually requires higher bias current and therefore higher power. The design must therefore balance gain, UGBW, power, and phase margin.}
\label{fig:schematics}
\end{figure}

\begin{figure*}[!h]
    \centering 
    \subfloat[GPT-4o-mini. Average and SEM of Top-10 FoM: \textbf{ACOF} $-0.359 \pm 0.018$, \textbf{LEDRO} $-0.643 \pm 0.015$, \textbf{Pure-BO} $-0.652 \pm 0.026$.\label{fig:fom_gpt4mini}]{
        \includegraphics[width=0.45\linewidth, trim=5 6 15 8.25mm, clip]{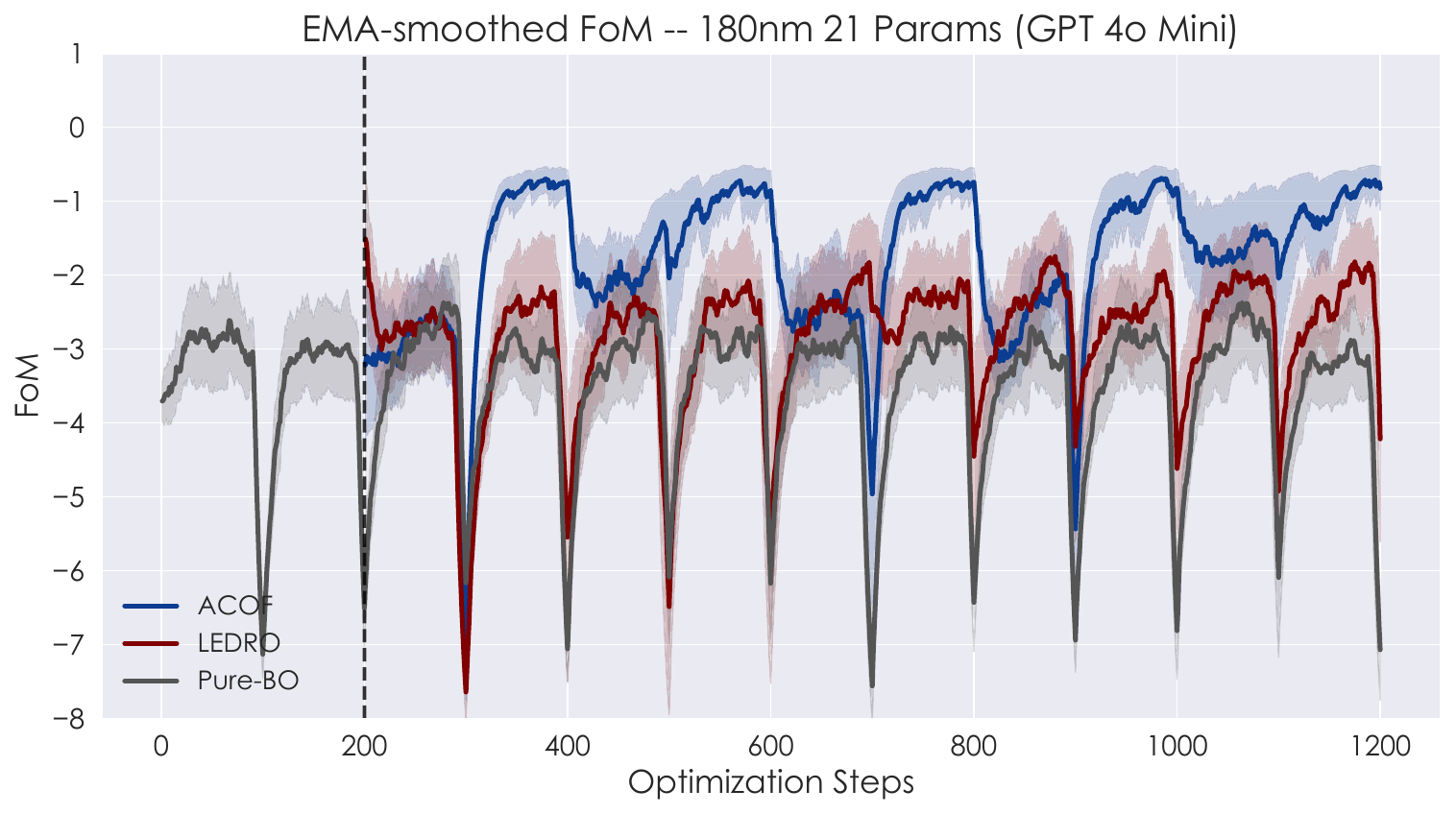}
    }\hfill
    \subfloat[GPT-5. Average and SEM of Top-10 FoM: \textbf{ACOF} $-0.271 \pm 0.024$, \textbf{LEDRO} $-0.319 \pm 0.035$, \textbf{Pure-BO} $-0.639 \pm 0.019$.\label{fig:fom_gpt5}]{
        \includegraphics[width=0.45\linewidth, trim=5 6 15 8.25mm, clip]{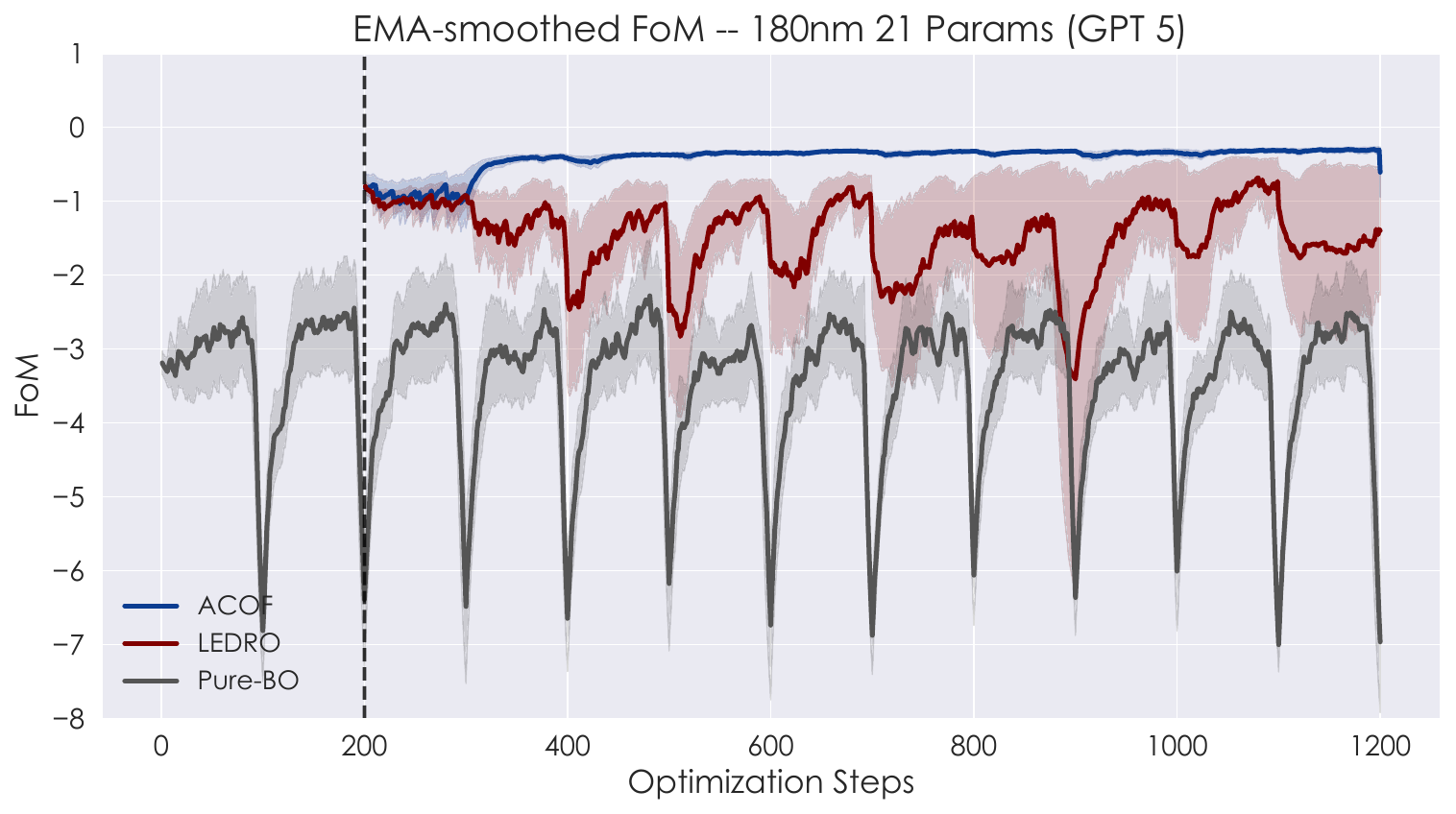}
    }
    \caption{Exponentially-smoothed FoM trajectories for the 180nm 21-parameter folded-cascode benchmark using variants of GPT as the LLM. Subcaptions report the across-run mean of the per-run top-10 mean FoM. Optimization steps for the LLM-based methods begin at 200 because both are initialized with $\mathcal{C}_0 = 200$ seed designs generated by Pure-BO. After initialization, the LLM-based methods update the optimization ranges every 100 steps. }
    \label{fig:fom_gpt}
\end{figure*}



\subsection{Benchmarks}

We benchmark four circuits of varying complexity across two PDKs. In SKY130 (130\,nm), we consider a two-stage amplifier and a cascode wide-swing amplifier with design vectors $x \in \mathbb{R}^{12}$ and $x \in \mathbb{R}^{17}$, respectively. In GF180MCU (180\,nm), we consider a two-stage amplifier and a folded-cascode single-ended amplifier with design vectors $x \in \mathbb{R}^{12}$ and $x \in \mathbb{R}^{21}$, respectively (see Fig. \ref{fig:schematics}). We selected these circuits due to their high real-world usage. Each design vector includes the tunable devices, bias, and compensation parameters. Our tests span $12-21$ parameters to show scalability across the parameter range. Note that our experimental breadth exceeds that of prior art published at top venues: ADO-LLM \cite{yin2024ado} tests two circuits on one PDK, and AnaFlow \cite{ahmadzadeh2025anaflow} tests two opamps on one PDK. 


\subsection{Implementations and Baselines}

The first setup employs the proposed {ACOF}. For comparison, we use a second setup that replaces the actor-critic loop with a single LLM coupled to the same BO backend, and a third setup that uses BO alone with no LLM  guidance. All three methods follow the same round schedule and the same per-round simulation budget under identical circuit-simulation conditions. In \textbf{ACOF}, the actor and critic are prompted separately. The actor is given topology-specific context, and the calibration set from the first round or a packed summary of the strongest valid designs from the previous round, including gain, UGBW, phase margin, power, and operating-region information. It then proposes numeric subranges for all tunable parameters. The critic receives those proposed ranges and returns a structured audit that either accepts the actor ranges or repairs them minimally within the legal bounds before BO is launched. After BO evaluates the candidate designs in the critic-approved region, the round summary is written back into the next actor prompt so that the search region can be revised progressively rather than reinitialized from scratch. See Fig. \ref{fig:convo} for an example.  \textbf{For clarity and reproducibility, we will release the full prompting templates upon acceptance.}

Our state-of-the-art comparison is a reimplementation of \textbf{LEDRO} \cite{kochar2025ledro}, modulating only the backbone language model to interface with BO. As an LLM-free comparison, we report the performance of \textbf{Pure-BO}, which optimizes the design using BO, \cite{wang_bayesopt_git}. Before the rounds begin, an initial set of designs is generated using BO to provide $\mathcal{C}_0 = 200$ starting points, which are used to condition the LLM baselines. In each round, we perform the optimization procedures and use Ngspice to compute the FoMs \cite{ngspice_v45_manual}. In our experiments, we configure our LLM-based methods using the open-source Qwen2.5-14B-Instruct \cite{qwen25_14b_instruct_modelcard}. To see how performance changes with respect to different LLM capabilities, we rerun our experiments on the 21-parameter folded-cascode benchmark using GPT-4o-mini \cite{gpt4} and GPT-5 \cite{gpt5}.

\subsection{Performance Metrics}

\noindent\textbf{Design Quality.} We measure and maximize FoM computed from gain $G(x)$, unity-gain bandwidth $BW(x)$, phase margin $\phi(x)$, and power $W(x)$. We define circuit-specific targets $(G^*, BW^*, \phi^*, W^*)$, reported as \texttt{Target Spec.} in Table.~\ref{tab:qwen_super_table}, and use them to construct normalized component scores $s_G(x)$, $s_{BW}(x)$, $s_{\phi}(x)$, and $s_W(x)$. Each score is zero when its corresponding target is met and penalizes deviations otherwise; we then combine them as $f(x)=3\cdot s_G(x)+s_{BW}(x)+s_{\phi}(x)+s_W(x)$. These quantities are not reported from a single best design. Instead, for each run and method, we sort all simulated designs by FoM and average each metric over the top-10 designs in our experiments. This gives a more stable summary of the quality of the best part of the search than reporting only a single extreme point.

\noindent\textbf{Reliability.} We report two run-level rates. \textit{Sim.\ Valid} is the fraction of all attempted designs in a run whose simulator output was marked valid by the pipeline. \textit{Phys.\ Feasible} is the fraction of attempted designs that satisfy our physical sanity checks. In our implementation, a design is counted as physically meaningful only if it yields positive bandwidth, positive phase margin, positive power, and nonnegative gain. Both reliability metrics are computed over the full set of attempts.

\begin{table}[hb]
\caption{Results on the 180nm 21-parameter folded-cascode benchmark using variants of GPT as the LLM. }
\label{tab:gpt_table}
\centering
\setlength{\tabcolsep}{4pt}
\renewcommand{\arraystretch}{1.15}
\resizebox{\columnwidth}{!}{%
\begin{tabular}{llccc}
\toprule
Model & Method & Top-10 FoM $\uparrow$ & Phys. Feasible (\%) $\uparrow$ & Regret $\downarrow$ \\
\midrule
\multirow{2}{*}{GPT-4o Mini} & ACOF & \textbf{-0.359$_{0.018}$} & \textbf{67.1$_{3.5}$} & \textbf{0.395$_{0.019}$} \\
 & LEDRO & -0.643$_{0.015}$ & 37.8$_{2.3}$ & 0.579$_{0.021}$ \\
\midrule
\multirow{2}{*}{GPT-5} & ACOF & \textbf{-0.271$_{0.024}$} & \textbf{99.4$_{0.0}$} & \textbf{0.313$_{0.019}$} \\
 & LEDRO & -0.319$_{0.035}$ & 80.9$_{8.9}$ & 0.381$_{0.024}$ \\
\bottomrule
\end{tabular}%
}
\end{table}

\noindent\textbf{Exploration/Exploitation.} To summarize how broadly each method visits distinct parts of the design space, we report \textit{Regions}. For each circuit, we first normalize the active design parameters to $[0,1]$ using that circuit's own parameter ranges. We then pool all sampled designs from a given run and method, and apply HDBSCAN \cite{hdbscan}, an unsupervised clustering method, in the normalized design space. The reported value is the number of discovered regions. A larger value indicates that the method visited more distinct regions of the parameter space during the run. This metric is computed from the full run-level point cloud. We also report \textit{Regret}, defined relative to the ideal FoM value of $0$. For each run and method, at each step we compute the gap between $0$ and the best FoM found up to that point, and report the average of that quantity over the run. Lower regret means that the method reached strong designs earlier and stayed closer to the ideal score throughout the run.


\section{Discussion and Conclusion} \label{sec:discussion}

 Across all four benchmarks in Table~\ref{tab:qwen_super_table} and our comparisons in Table \ref{tab:gpt_table}, ACOF performs best in terms of FoM and regret. This pattern is more informative than any single raw specification because the tables average over the top-10 designs, reflecting the quality of a region rather than a single fortunate sample. The baselines occasionally lead on isolated metrics--Pure-BO on phase margin in some cases and on bandwidth in two circuits, and LEDRO on gain for the 180\,nm 21-parameter benchmark--but those wins do not carry through to the joint objective. ACOF repeatedly converges to design sets in which gain, bandwidth, stability, and power sit in a better overall balance.

The benefit is not tied to a particular language model (Fig. \ref{fig:fom_gpt}). ACOF rises to stronger FoM levels and remains there through most of the search. Using GPT-4o-mini (Fig.~\ref{fig:fom_gpt4mini}), ACOF is 44.2\% closer to the ideal FoM of 0 than LEDRO is. Both ACOF and LEDRO performed better when using the stronger GPT-5 (Fig.~\ref{fig:fom_gpt5}). However, ACOF still outperformed LEDRO. This result shows that the framework is not tied to one particular LM and that a better underlying model can be converted into better optimization behavior without changing the search procedure.

The reliability columns in Table~\ref{tab:qwen_super_table} show that ACOF stays at or near perfect simulator validity across the Qwen experiments. That same pattern carries over to physical feasibility in Table~\ref{tab:gpt_table} on the 180\,nm 21-parameter folded-cascode benchmark with GPT-4o Mini and GPT-5. In both GPT settings, ACOF achieves higher physical feasibility than LEDRO while also improving top-10 FoM and regret. On the corresponding Qwen benchmark, nearly every ACOF proposal simulates successfully, and a much larger fraction remains physically meaningful than for LEDRO or Pure-BO. Similar feasibility gains also appear on the 130\,nm 12-parameter and 180\,nm 12-parameter circuits. Viewed alongside the actor-critic exchange in Fig.~\ref{fig:convo}, the pattern suggests that the critic steers the search toward promising designs. In practice, more of the simulation budget is spent on candidates that are admissible and useful.

The exploration results show that ACOF's movement across regions is productive rather than merely broad. This is visible both in the region counts in Table~\ref{tab:qwen_super_table} and in the embedding trajectories in Fig.~\ref{fig:umap_gpt4mini}. On the 180\,nm 21-parameter circuit, the red contours for ACOF migrate across several separated pockets of the embedding over successive round pairs, including pockets at $\texttt{UMAP Component 2} > 1$ that the baselines touch only lightly. What matters is that these relocations continue to pay off: in every round pair, the ACOF panels report a better average FoM than LEDRO and Pure-BO, and the best FoM found within each pair is stronger as well. Table~\ref{tab:qwen_super_table} quantifies the same effect. On this benchmark, ACOF covers the largest number of regions and still records the best FoM and regret. A similar relation between region coverage and solution quality appears on the 180\,nm 12-parameter and 130\,nm 17-parameter circuits, where ACOF again visits more regions than either baseline. The 130\,nm 12-parameter case shows the contrast from the other side: Pure-BO touches the most regions there, yet its physical-feasibility rate drops to $26.9\%$. The distinction, then, is not movement alone, but movement that lands in workable territory.

Taken together, Table~\ref{tab:qwen_super_table}, Table~\ref{tab:gpt_table}, Fig.~\ref{fig:umap_gpt4mini}, and Fig.~\ref{fig:fom_gpt} support that ACOF gives a stronger search process for analog sizing. This effect is clearest on the 180\,nm 21-parameter benchmark, where the larger search space and tighter metric coupling make selective exploration especially valuable. These findings position ACOF as a robust and sample-efficient framework for analog design automation.




\balance
\bibliographystyle{ACM-Reference-Format}
\bibliography{allBib1}      

@article{lyu2017efficient,
  title={An efficient Bayesian optimization approach for automated optimization of analog circuits},
  author={Lyu, Wenlong and Xue, Pan and Yang, Fan and Yan, Changhao and Hong, Zhiliang and Zeng, Xuan and Zhou, Dian},
  journal={IEEE Transactions on Circuits and Systems I: Regular Papers},
  volume={65},
  number={6},
  pages={1954--1967},
  year={2017},
  publisher={IEEE}
}

@inproceedings{touloupas2021local,
  title={Local Bayesian optimization for analog circuit sizing},
  author={Touloupas, Konstantinos and Chouridis, Nikos and Sotiriadis, Paul P},
  booktitle={2021 58th ACM/IEEE design automation conference (DAC)},
  pages={1237--1242},
  year={2021},
  organization={IEEE}
}

@inproceedings{gu2024tss,
  title={tSS-BO: Scalable Bayesian Optimization for Analog Circuit Sizing via Truncated Subspace Sampling},
  author={Gu, Tianchen and Wang, Jiaqi and Bi, Zhaori and Yan, Changhao and Yang, Fan and Qin, Yajie and Cui, Tao and Zeng, Xuan},
  booktitle={2024 Design, Automation \& Test in Europe Conference \& Exhibition (DATE)},
  pages={1--6},
  year={2024},
  organization={IEEE}
}

@article{liu2025layoutcopilot,
  title={Layoutcopilot: An llm-powered multi-agent collaborative framework for interactive analog layout design},
  author={Liu, Bingyang and Zhang, Haoyi and Gao, Xiaohan and Kong, Zichen and Tang, Xiyuan and Lin, Yibo and Wang, Runsheng and Huang, Ru},
  journal={IEEE Transactions on Computer-Aided Design of Integrated Circuits and Systems},
  year={2025},
  publisher={IEEE}
}

@article{chen2024llm,
  title={LLM-enhanced Bayesian optimization for efficient analog layout constraint generation},
  author={Chen, Guojin and Zhu, Keren and Kim, Seunggeun and Zhu, Hanqing and Lai, Yao and Yu, Bei and Pan, David Z},
  journal={arXiv preprint arXiv:2406.05250},
  year={2024}
}

@article{liu2024ampagent,
  title={Ampagent: An llm-based multi-agent system for multi-stage amplifier schematic design from literature for process and performance porting},
  author={Liu, Chengjie and Chen, Weiyu and Peng, Anlan and Du, Yuan and Du, Li and Yang, Jun},
  journal={arXiv preprint arXiv:2409.14739},
  year={2024}
}

@inproceedings{lai2025analogcoder,
  title={Analogcoder: Analog circuit design via training-free code generation},
  author={Lai, Yao and Lee, Sungyoung and Chen, Guojin and Poddar, Souradip and Hu, Mengkang and Pan, David Z and Luo, Ping},
  booktitle={Proceedings of the AAAI Conference on Artificial Intelligence},
  volume={39},
  number={1},
  pages={379--387},
  year={2025}
}

@inproceedings{yin2024ado,
  title={Ado-llm: Analog design bayesian optimization with in-context learning of large language models},
  author={Yin, Yuxuan and Wang, Yu and Xu, Boxun and Li, Peng},
  booktitle={Proceedings of the 43rd IEEE/ACM International Conference on Computer-Aided Design},
  pages={1--9},
  year={2024}
}

@inproceedings{kochar2025ledro,
  title={Ledro: Llm-enhanced design space reduction and optimization for analog circuits},
  author={Kochar, Dimple Vijay and Wang, Hanrui and Chandrakasan, Anantha P and Zhang, Xin},
  booktitle={2025 IEEE International Conference on LLM-Aided Design (ICLAD)},
  pages={141--148},
  year={2025},
  organization={IEEE}
}

@article{shinn2023reflexion,
  title={Reflexion: Language agents with verbal reinforcement learning},
  author={Shinn, Noah and Cassano, Federico and Gopinath, Ashwin and Narasimhan, Karthik and Yao, Shunyu},
  journal={Advances in Neural Information Processing Systems},
  volume={36},
  pages={8634--8652},
  year={2023}
}

@INPROCEEDINGS{lorakd,
  author={Rouf, Nirjhor and Amin, Fin and Franzon, Paul D.},
  booktitle={2024 IEEE LLM Aided Design Workshop (LAD)}, 
  title={Can Low-Rank Knowledge Distillation in LLMs be Useful for Microelectronic Reasoning?}, 
  year={2024},
  volume={},
  number={},
  pages={1-6},
  doi={10.1109/LAD62341.2024.10691755}}

@manual{ngspice_v45_manual,
  title        = {Ngspice User’s Manual},
  author       = {Vogt, Holger and Atkinson, Giles and Nenzi, Paolo},
  year         = {2025},
  month        = sep,
  note         = {Version 45 (ngspice release version)},
  url          = {https://ngspice.sourceforge.io/docs/ngspice-45-manual.pdf},
}

@book{barros2010analog,
  title={Analog circuits and systems optimization based on evolutionary computation techniques},
  author={Barros, Manuel FM and Guilherme, Jorge MC and Horta, Nuno CG},
  volume={9},
  year={2010},
  publisher={Springer}
}

@inproceedings{ahmadzadeh2024using,
  title={Using probabilistic model rollouts to boost the sample efficiency of reinforcement learning for automated analog circuit sizing},
  author={Ahmadzadeh, Mohsen and Gielen, Georges GE},
  booktitle={Proceedings of the 61st ACM/IEEE Design Automation Conference},
  pages={1--6},
  year={2024}
}

@article{liu2025eesizer,
  title={Eesizer: Llm-based ai agent for sizing of analog and mixed signal circuit},
  author={Liu, Chang and Chitnis, Danial},
  journal={IEEE Transactions on Circuits and Systems I: Regular Papers},
  year={2025},
  publisher={IEEE}
}

@article{somayaji2025llm,
  title={LLM-USO: Large Language Model-based Universal Sizing Optimizer},
  author={Somayaji, NS Karthik and Li, Peng},
  journal={IEEE Transactions on Computer-Aided Design of Integrated Circuits and Systems},
  year={2025},
  publisher={IEEE}
}

@inproceedings{ahmadzadeh2025anaflow,
  title={AnaFlow: Agentic LLM-based workflow for reasoning-driven explainable and sample-efficient analog circuit sizing},
  author={Ahmadzadeh, Mohsen and Chen, Kaichang and Gielen, Georges},
  booktitle={2025 IEEE/ACM International Conference On Computer Aided Design (ICCAD)},
  pages={1--7},
  year={2025},
  organization={IEEE}
}

@article{nelson2024needle,
  title={Needle in the haystack for memory based large language models},
  author={Nelson, Elliot and Kollias, Georgios and Das, Payel and Chaudhury, Subhajit and Dan, Soham},
  journal={arXiv preprint arXiv:2407.01437},
  year={2024}
}

@book{sutton2018reinforcement,
  title     = {Reinforcement Learning: An Introduction},
  author    = {Sutton, Richard S. and Barto, Andrew G.},
  edition   = {2},
  year      = {2018},
  publisher = {MIT Press},
  address   = {Cambridge, MA},
  url       = {https://incompleteideas.net/book/the-book-2nd.html}
}

@inproceedings{sutton2000policy,
  title     = {Policy Gradient Methods for Reinforcement Learning with Function Approximation},
  author    = {Sutton, Richard S. and McAllester, David A. and Singh, Satinder P. and Mansour, Yishay},
  booktitle = {Advances in Neural Information Processing Systems 12 (NeurIPS 1999)},
  editor    = {S. A. Solla and T. K. Leen and K.-R. M{\"u}ller},
  pages     = {1057--1063},
  year      = {2000},
  publisher = {MIT Press},
}

@article{mnih2015human,
  title   = {Human-level Control through Deep Reinforcement Learning},
  author  = {Mnih, Volodymyr and Kavukcuoglu, Koray and Silver, David and Rusu, Andrei A. and Veness, Joel and Bellemare, Marc G. and Graves, Alex and Riedmiller, Martin and Fidjeland, Andreas K. and Ostrovski, Georg and Petersen, Stig and Beattie, Charles and Sadik, Amir and Antonoglou, Ioannis and King, Helen and Kumaran, Dharshan and Wierstra, Daan and Legg, Shane and Hassabis, Demis},
  journal = {Nature},
  volume  = {518},
  number  = {7540},
  pages   = {529--533},
  year    = {2015},
  doi     = {10.1038/nature14236},
  url     = {https://www.nature.com/articles/nature14236}
}

@inproceedings{schulman2015trpo,
  title     = {Trust Region Policy Optimization},
  author    = {Schulman, John and Levine, Sergey and Abbeel, Pieter and Jordan, Michael and Moritz, Philipp},
  booktitle = {Proceedings of the 32nd International Conference on Machine Learning},
  series    = {Proceedings of Machine Learning Research},
  volume    = {37},
  pages     = {1889--1897},
  year      = {2015},
  publisher = {PMLR},
  url       = {https://proceedings.mlr.press/v37/schulman15.html}
}

@inproceedings{haarnoja2018sac,
  title     = {Soft Actor-Critic: Off-Policy Maximum Entropy Deep Reinforcement Learning with a Stochastic Actor},
  author    = {Haarnoja, Tuomas and Zhou, Aurick and Abbeel, Pieter and Levine, Sergey},
  booktitle = {Proceedings of the 35th International Conference on Machine Learning},
  series    = {Proceedings of Machine Learning Research},
  volume    = {80},
  pages     = {1861--1870},
  year      = {2018},
  publisher = {PMLR},
  url       = {https://proceedings.mlr.press/v80/haarnoja18b.html}
}

@article{schrittwieser2020muzero,
  title   = {Mastering Atari, Go, Chess and Shogi by Planning with a Learned Model},
  author  = {Schrittwieser, Julian and Antonoglou, Ioannis and Hubert, Thomas and Simonyan, Karen and Sifre, Laurent and Schmitt, Simon and Guez, Arthur and Lockhart, Edward and Hassabis, Demis and Graepel, Thore and Lillicrap, Timothy and Silver, David},
  journal = {Nature},
  volume  = {588},
  number  = {7839},
  pages   = {604--609},
  year    = {2020},
  doi     = {10.1038/s41586-020-03051-4},
  url     = {https://www.nature.com/articles/s41586-020-03051-4}
}

@inproceedings{christiano2017preferences,
  title     = {Deep Reinforcement Learning from Human Preferences},
  author    = {Christiano, Paul F. and Leike, Jan and Brown, Tom B. and Martic, Miljan and Legg, Shane and Amodei, Dario},
  booktitle = {Advances in Neural Information Processing Systems 30 (NeurIPS 2017)},
  editor    = {I. Guyon and U. V. Luxburg and S. Bengio and H. Wallach and R. Fergus and S. Vishwanathan and R. Garnett},
  pages     = {4299--4307},
  year      = {2017},
  publisher = {Curran Associates, Inc.},
  url       = {https://papers.neurips.cc/paper/7017-deep-reinforcement-learning-from-human-preferences.pdf}
}

@inproceedings{ouyang2022instructgpt,
  title     = {Training Language Models to Follow Instructions with Human Feedback},
  author    = {Ouyang, Long and Wu, Jeff and Jiang, Xu and Almeida, Diogo and Wainwright, Carroll L. and Mishkin, Pamela and Zhang, Chong and Agarwal, Sandhini and Slama, Katarina and Ray, Alex and Schulman, John and Hilton, Jacob and Kelton, Fraser and Miller, Luke and Simens, Maddie and Askell, Amanda and Welinder, Peter and Christiano, Paul F. and Leike, Jan and Lowe, Ryan},
  booktitle = {Advances in Neural Information Processing Systems 35 (NeurIPS 2022)},
  editor    = {S. Koyejo and S. Mohamed and A. Agarwal and D. Belgrave and K. Cho and A. Oh},
  pages     = {27730--27744},
  year      = {2022},
  publisher = {Curran Associates, Inc.},
  url       = {https://proceedings.neurips.cc/paper_files/paper/2022/file/b1efde53be364a73914f58805a001731-Paper-Conference.pdf}
}

@misc{wang_bayesopt_git,
  author       = {Hao Wang},
  title        = {Bayesian-Optimization},
  howpublished = {\url{https://github.com/wangronin/Bayesian-Optimization}},
  year         = {2023},
  note         = {GitHub repository},
}

@misc{qwen25_14b_instruct_modelcard,
  author       = {{Qwen Team}},
  title        = {Qwen2.5-14B-Instruct},
  year         = {2024},
  howpublished = {\url{https://huggingface.co/Qwen/Qwen2.5-14B-Instruct}},
  note         = {Model card, accessed: 2026-03-15}
}

@article{hdbscan,
  title={hdbscan: Hierarchical density based clustering.},
  author={McInnes, Leland and Healy, John and Astels, Steve and others},
  journal={J. Open Source Softw.},
  volume={2},
  number={11},
  pages={205},
  year={2017}
}

@article{gpt5,
  title={Openai gpt-5 system card},
  author={Singh, Aaditya and Fry, Adam and Perelman, Adam and Tart, Adam and Ganesh, Adi and El-Kishky, Ahmed and McLaughlin, Aidan and Low, Aiden and Ostrow, AJ and Ananthram, Akhila and others},
  journal={arXiv preprint arXiv:2601.03267},
  year={2025}
}

@article{gpt4,
  title={Gpt-4o system card},
  author={Hurst, Aaron and Lerer, Adam and Goucher, Adam P and Perelman, Adam and Ramesh, Aditya and Clark, Aidan and Ostrow, AJ and Welihinda, Akila and Hayes, Alan and Radford, Alec and others},
  journal={arXiv preprint arXiv:2410.21276},
  year={2024}
}

@inproceedings{wen2022high,
  title={High dimensional optimization for electronic design},
  author={Wen, Yuejiang and Dean, Jacob and Floyd, Brian A and Franzon, Paul D},
  booktitle={Proceedings of the 2022 ACM/IEEE Workshop on Machine Learning for CAD},
  pages={153--157},
  year={2022}
}

@inproceedings{TD3,
  title     = {Addressing Function Approximation Error in Actor-Critic Methods},
  author    = {Fujimoto, Scott and van Hoof, Herke and Meger, David},
  booktitle = {Proceedings of the 35th International Conference on Machine Learning (ICML)},
  pages     = {1587--1596},
  year      = {2018}
}

@article{ahmadzadeh2025anacraft,
  title={AnaCraft: Duel-play probabilistic-model-based reinforcement learning for sample-efficient PVT-robust analog circuit sizing optimization},
  author={Ahmadzadeh, Mohsen and Lappas, Jan and Wehn, Norbert and Gielen, Georges},
  journal={IEEE Transactions on Computer-Aided Design of Integrated Circuits and Systems},
  year={2025},
  publisher={IEEE}
}

\end{document}